# Natural Disaster Classification using Aerial Photography Explainable for Typhoon Damaged Feature

Yasuno Takato[1], Amakata Masazumi[1], Okano Masahiro[1]

[1] Research Institute for Infrastructure Paradigm Shift, Yachiyo Engineering Co., Ltd.,
Asakusabashi 5-20-8, Taito-ku, Tokyo, Japan
`{tk-yasuno,amakata,ms-okano}@yachiyo-eng.co.jp`

**Abstract.** Recent years, typhoon damages has become social problem owing to climate change. In 9 September 2019, Typhoon Faxai passed on the Chiba in Japan, whose damages included with electric provision stop because of strong wind recorded on the maximum 45 meter per second. A large amount of tree fell down, and the neighbour electric poles also fell down at the same time. These disaster features have caused that it took 18 days for recovery longer than past ones. Immediate responses are important for faster recovery. As long as we can, aerial survey for global screening of devastated region would be required for decision support to respond where to recover ahead. This paper proposes a practical method to visualize the damaged areas focused on the typhoon disaster features using aerial photography. This method can classify eight classes which contains land covers without damages and areas with disaster. Using target feature class probabilities, we can visualize disaster feature map to scale a colour range. Furthermore, we can realize explainable map on each unit grid images to compute the convolutional activation map using Grad-CAM. We demonstrate case studies applied to aerial photographs recorded at the Chiba region after typhoon.

## 1 Introduction

### 1.1 Typhoon Damage Prediction for Immediate Response and Recovery

This paper highlights the typhoon disaster feature extracted by the aero-photographs because it is reasonable cost more than satellite imagery and it has advantage of global sky view more than drone images. As shown these related works at next section 1.2, one of typhoon disaster features, whose complex features contains the tree-fallen, is not clearly explored. First, the tree-fallen feature has green-colour of leafs and blown-colour of blanches, so this region of interest is completely colour segmented using the straight forward colour slicing and the standard image processing. Second, it seems that the shape is various owing to the damage intensity of tree-fallen range from light level to extremely one. More flexible learning methods are required for disaster features extraction using deep neural networks.

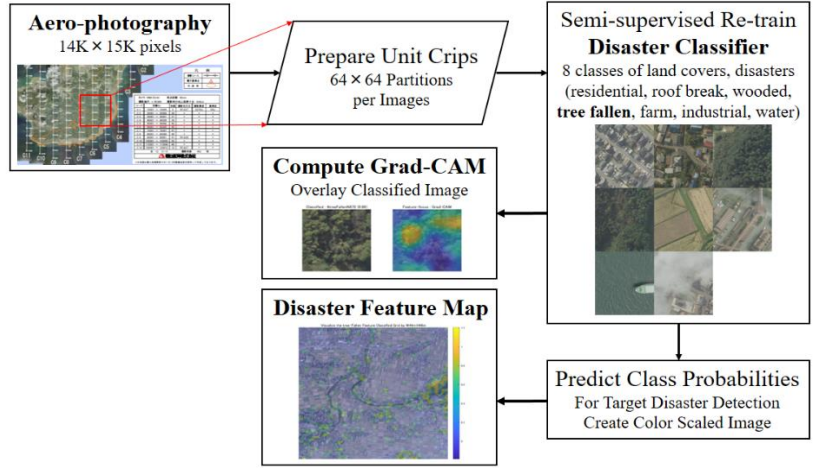

**Fig. 1.** A proposed workflow to classify disaster damages and to visualize a target feature

Fig.1 shows our proposed workflow to classify disaster damages and to visualize a target classified feature. Immediate responses towards typhoon damages are important for faster recovery which consists with electric stop caused by tree-fallen. Therefore, more reasonable sky view screening of devastated region is required for decision support to respond where to prioritize recovery actions. An advantage of aerial photograph is higher resolution that is 0.2m, than satellite imagery and it also enables to wider search efficiently than drone flight. This paper proposes a method to predict the damaged area focused on the typhoon disaster features using aero-photograph images.

## 1.2 Related works and papers

**Aerial Survey for Natural Disaster Monitoring.** In order to recognize a status after natural disaster, several aerial surveys are available from various height overview such as satellite imagery, aircraft, helicopter, and drone. There are many related works around the devastated area where some disaster occurred at each phase of preparedness, urgent response, and recovery. For example, Chou et al.[1] proposed the drone-based photographing workflow for disaster monitoring and management operation to get real-time aerial photos. The pro of drone use is to overcome transport barriers to quickly acquire all the major affected area of the real-time disaster information. But, the con is that drone is not able to keep a long flight more than 30 minutes. Kentosch et al [2] studied UAV based tree species classification task in Japanese mixed forests for deep learning in two different mixed forest types. This research has tended to focus on the identification tree species rather than the natural disaster feature such as tree fallen. Next, the JICA survey team [3] used a helicopter owned by the air force of Sri Lanka to investigate the flood and sediment disaster damages caused by the strong winds and the heavy rainfalls during 15 to 18 May 2016, named as Cyclone Roanu. They also collected the Light Detection and Ranging (LiDAR) data to compare between before normal status and after landslide. The aerial photograph could visualize the damages where roads are flooded, river water exceeded a dike in residential areas, landslide or collapse, debris flow on the top of mountain. The advantage of LiDAR is that beside rain and clouds, no other weather conditions limit their use [4]. As an active sensor, missions in the night-time are possible. In Japan, there is a Bosai platform [5] for disaster preparedness, response and recovery. Several private company have investigated after natural disaster using aerial survey and made a disaster mapping for emergency use. However, it takes high cost of satellite image over global viewpoint and it needs longer time to get a target image of fully covered region. On the other hand, drone surveillance takes lowest cost though the flight time is twenty or thirty minutes, and the scope is narrow due to the constraint of height available flight.

## 1.3   Feature Extraction for Natural Disaster Damage Assessment

There are a lot of earthquake disaster detection especially building damage using satellite imagery, LiDAR point clouds, and drone images. For example, He et al. [6] presented 3D shape descriptor of individual roofs for detecting surface damage and roof exhibiting structural damage. However, it is not always possible to acquire an airborne LiDAR data of devastated region to be full-covered after each disaster. This approach would take too high cost to prepare the input data source for damage assessment. In addition, Nex et al. [7] presented a method to autonomously map building damages with a drone in near real-time. Unfortunately, this study is too narrow experience to explore the practical use for disaster response. The limitation of method based on drone is also the short flight time less than 30 minutes to search the devastated region. Regarding typhoon damage assessment, Liu et al. [8] reported the Landsat-8 automatic image processing system (L-8 AIPS) which enables to the most up-to-date scenes of Taiwan to be browsed, and in only one hour's time after receiving the raw data from United States Geological Survey (USGS) level-1 product. However, this application depended on the specific satellite imagery and each country must customize such an application, the limitation is the 15m resolution for monitoring disaster damages. Furthermore, Gupta et al [9] proposed to identify impacted areas and accessible roads and buildings in post-disaster scenarios. However, this research has been tended to focus on the grasp of post-disaster recovery scenario rather than the immediate respond of early restoration. On the other hand, Rahnemoonfar et al [10] studied a hurricane Michael dataset HRUD for visual perception in disaster scenarios using semantic segmentation including with the class of building destruction. But, one of the major drawbacks to use the semantic segmentation is the scarcity of natural disaster feature, the difficulty of data mining for high accuracy, and much workforce of annotation for accurate damage region of interest. Furthermore, Sheykhmousa et al. [11] focused on the post-disaster recovery process before typhoon and after four years, and they classified the five classes of land cover and land use changes to visualize the recovery map which enables to signal positive recovery, neutral, and negative recovery, etc. Natural forces are heavy winds, storm surges, and destroyed existing urban facilities. Their damage classes contained with inundated land and flattened trees. They demonstrated a land cover and land use recovery map that quantify the post-disaster recovery process at the pixel level. However, this study is not immediate recovery but four years long recovery process. The classified recovery map is too far view based on satellite imagery at the 2m resolution to extract the damage feature in more detail.

## 2 Modelling

### 2.1 Partition Clips and Learning Disaster Features

**Prepare 64 by 64 Partitioned Clips Divided 14K by 15K pixels into 221 by 243.**
Fig.2 shows a proposed machine learning workflow that consist of the former part of disaster damaged annotation and the later part of classification modelling. First, the size of aero-photographs is 14 thousands by 15 thousands pixels, so the original size is too large and it is not fitted as the input of deep neural networks whose size is frequently 224 by 224 and 229 by 229. High quality learning requires to keep the original RGB data per pixel without resize as far as we can. Therefore, this method make a set of base partitions 64 by 64 into each unit grid image clips with 221 by 243 pixel size. This aerial photograph has the 0.196m resolution so that the real size of unit grid square stands for 44 by 48 meter distance.

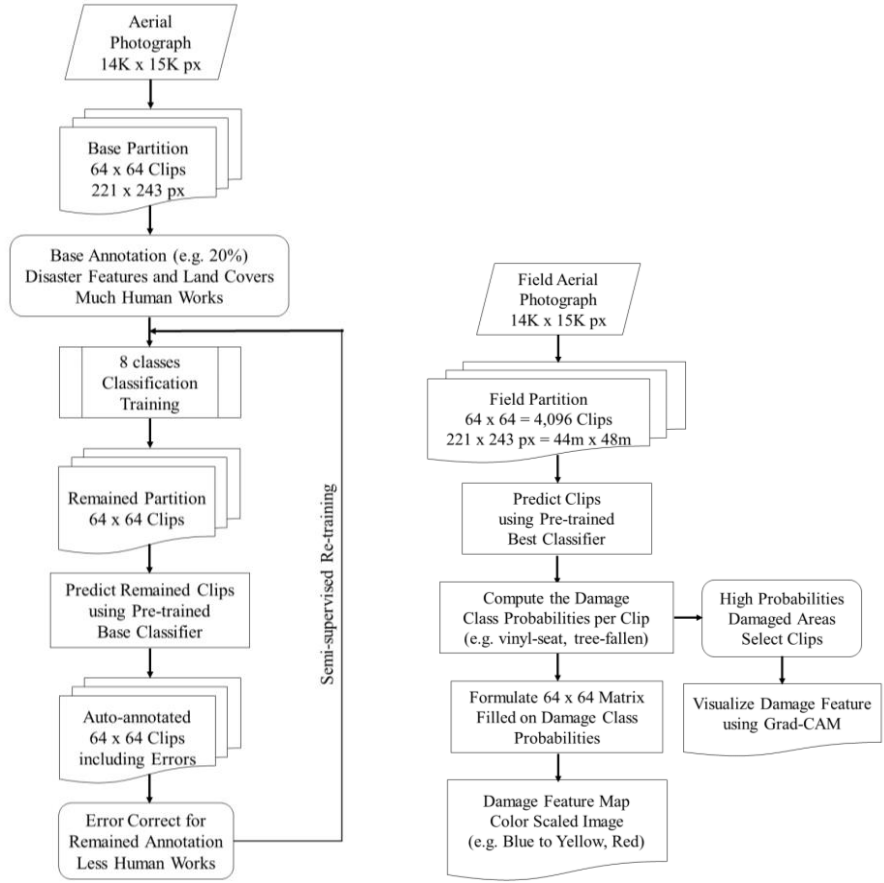

**Fig. 2.** (Left) A Machine Learning workflow of Disaster Damaged Annotation and Classification, (Right) An application workflow for Damaged Feature Mapping Filled on Class Probabilities and for Visualized Damage Feature per Clip Area computing the Grad-CAM.

**Table 1.** Hierarchical Classes Setting including Land Covers and Typhoon Disaster Features

| 1 residential area (without damage) | 5 farm field (rice, crop) |
|---|---|
| 2 **house roof break** (remained vinyl seat) | 6 industrial area (manufacturing) |
| 3 wooded field (without damage) | 7 water cover (sea, river, lake) |
| 4 **tree fallen field** (caused by strong wind) | 8 cloudy cover (partial or full) |

**Efficient Annotation via Semi-supervised Prediction using Pre-trained Classifier.**
This paper highlights on the typhoon disaster feature such as tree-fallen and normal land covers without damages. Table 1 shows a proposed hierarchical eight classes which contains seven land covers and the tree-fallen features. We could use 26 images of aerial photograph. Total amount of clip images is almost always very large more than hundred thousands, so human annotation by hands is difficult to divide subgroup of land cover and disaster feature classes. Owing to the hurdle of efficient learning, firstly, we propose to train a base classifier using twenty percent of clip images with the order of twenty thousand images. Secondly, using remained clips with order of eighty thousands as the added input for semi-supervised learning, we can predict the label per clip and it takes less works of annotation because almost labels are correct and error corrections are less than the case without pre-trained model. And then using the classified labels, we can almost automate to divide their clips into eight subfolders efficiently. Thirdly, human check is possible whether each subfolder contains errors or not, and can correct their miss-classification of clips less works than fully human works.

**Accuracy Comparison with Deep Neural Networks.**
We can train classification models based on ten candidates of deep neural networks with advantage of the whole accuracy and two disaster classes of accuracies, that is house roof break and tree fallen. We propose and compare usable deep neural networks such as AlexNet [12,13], VGG16 [14], GoogleNet , Inception-v3 [15], ResNet50, ResNet101 [16], Inception-ResNet-v2 [17], DenseNet-201 [18]. Furthermore, MobileNet-v2 [19] and ShuffleNet [20,21] has less parameters and fitted to mobile devices, where the later network uses special techniques such as channel shuffle and group convolution. We can evaluate their accuracy with both recall and precision to compute their confusion matrix. Here, the recall is more important than precision because we should avoid miss-classification in spite of that it contains disaster features.

## 2.2 Visual Explanation toward Disaster Features

**Colour Scaled Feature Map using Target Class Probabilities.** Fig 2 shows an application workflow for damaged feature mapping filled on damaged class probabilities and for visualized damage feature per clip area computing the Grad-CAM [22]. Based on the already trained best classifier, we can compute target class probabilities per unit grid, and then we can formulate the damage feature matrix which consists 64 rows and 64 columns. Therefore, we enable to scale the matrix of specific class probabilities into colour feature map ranged among two colours [23]. This paper realize a damage feature map. We can formulate a tree fallen feature map to scale colour range from blue to yellow. The yellow side is huge damage, the blue side is small damage.

**Visual Disaster Explanation per Unit Grid using Grad-CAM.** In case that some grid area has a high probability at a specific damage class, we can create the micro damage feature map for more detailed scale per unit grid area. We can explain the damage features on each unit grid images to compute the convolutional activation map using Grad-CAM based on deep classification network layers with optimized parameters.

## 3   Applied results

### 3.1   Training and Test Dataset of Aerial-photographs

We have demonstrated training and test studies applied to aero-photographs with the size of 14 thousands pixels by 15 thousands ones recorded at the south Chiba region after 18 days at the post typhoon disaster, 27 to 28 September 2019, provided by Aero Asahi Co.Ltd. Here, the real size per pixel on the land is 19.6 cm so that the size of unit grid square contains 44 by 48 meter distance. The height to get aero-photographs is 3,270 meters and the number of them is 452 images. The number of clip images is 5,954 for a base classifier training at the region of Kimitsu city on the Chiba prefecture. Here, we set the division of train and test is 7 : 3. Several test aero-photographs are located at the region of Kyonan town and Tateyama city on the south Chiba prefecture.

**Table 2.** Accuracy comparison between various models

| CNN models | Total validation accuracy | house roof break (covered vinyl-seat) accuracy | | tree-fallen accuracy | | Frozen layer /Whole layer | Runing time (minutes) |
|---|---|---|---|---|---|---|---|
| | | recall | precision | recall | precision | | |
| AlexNet | 89.59 | 67.9 | 97.6 | 95.8 | 68.8 | 16/25 | 20 |
| VGG16 | 88.07 | 66.3 | 89.9 | 87.8 | 71.2 | 32/41 | 50 |
| GoogleNet | 89.75 | **78.6** | 92.7 | 93.1 | 69.5 | 110/144 | 27 |
| Inception-v3 | 90.26 | 70.1 | 97.7 | **99.9** | 66.1 | 249/315 | 79 |
| MobileNet-v2 | 88.63 | 64.6 | 98.1 | **99.6** | 65.6 | 104/154 | 38 |
| **ResNet50** | **90.48** | 74.5 | 96.8 | 99.6 | **74.1** | 120/177 | 46 |
| ResNet101 | 88.02 | 63.8 | **98.7** | 99.6 | 69.1 | 290/347 | 70 |
| **DenseNet-201** | **92.16** | 75.3 | 96.3 | 97.7 | **77.1** | 647/708 | 193 |
| Inception-ResNet-v2 | 88.52 | 69.5 | 98.3 | 95.1 | 74.3 | 766/824 | 175 |
| **ShuffleNet** | **93.39** | **82.3** | 98.1 | 98.9 | **79.7** | 137/172 | 33 |

### 3.2   Damage Feature Classifier Trained Results

The author applied to ten classification deep neural networks as base model towards the number of 5,954 of unit grid clips, which are randomly divided into rates with training 70 and 30 test. The training images are pre-processed by three specific augmentation operations such as randomly flip the training images along the vertical axis, and randomly translate them up to 30 pixels, and scale them up to 10% horizontally and vertically. Then transfer learning is useful for faster training and disaster response. Here, the author make freeze the round 70 to 80 percent of deep network layers at the concatenation point combined with convolutional layers and skip connections. Table 2 indicates the accuracy comparison between ten deep neural networks, which has each number of whole layers setting with the frozen layer point. In this study, the mini-batch size is 32, and the maximum epoch is 30 with shuffle every epoch. One epoch has 130 iterations, the total batch learning results in 3,900 iterations. Further, the initial learning rate is 0.0005, and learning rate schedule is every 5 epoch with drop factor 0.75. From computing results, the ShuffleNet has the most accurate for 30 percent test clips validation. Also the damage class accuracy of both recall and precision has higher score than other nine deep neural networks. Table 3 shows the confusion matrix of the most promising ShuffleNet as far as we implemented our comparison.

**Table 3.** Confusion matrix of the ShuffleNet best classifier (row : truth, column : prediction)

| | 1residencialN851 | 2vinylSeatN811 | 3woodedN752 | 4treeFallenN872 | 5farmN846 | 6industrialN691 | 7seaN517 | 8cloudyN614 | | |
|---|---|---|---|---|---|---|---|---|---|---|
| 1residencialN851 | 247 | 4 | | 4 | | | | | 96.9% | 3.1% |
| 2vinylSeatN811 | 15 | 200 | 4 | 16 | 8 | | | | 82.3% | 17.7% |
| 3woodedN752 | 1 | | 177 | 48 | | | | | 78.3% | 21.7% |
| 4treeFallenN872 | | | 3 | 259 | | | | | 98.9% | 1.1% |
| 5farmN846 | 5 | | 3 | 2 | 244 | | | | 96.1% | 3.9% |
| 6industrialN691 | | | | | | 207 | | | 100.0% | |
| 7seaN517 | | | | 4 | | | 151 | | 97.4% | 2.6% |
| 8cloudyN614 | 1 | | | | | | | 183 | 99.5% | 0.5% |
| | 91.8% | 98.0% | 92.7% | 79.7% | 95.3% | 100.0% | 100.0% | 100.0% | | |
| | 8.2% | 2.0% | 7.3% | 20.3% | 4.7% | | | | | |

### 3.3 Damage Feature Map and Unit Grid Visualization Results

**Tree-fallen feature mapping results.** Fig. 3 shows the tree-fallen damage feature map based on the class probabilities per clip. The colour is scaled between the blue and yellow. The yellow side is positive tree-fallen caused by strong wind. Here, we can understand that a lot of tree-fallen areas are located besides river and around mountain.

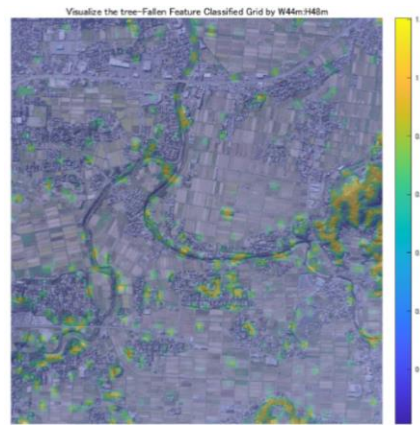

**Fig. 3.** Tree fallen feature map using damage class probabilities

Fig. 4 shows the visual explanation of tree-fallen more detail scale per unit grid with the size of 221×243 using Grad-CAM, where each pair of original clip and activation map are overlaid. The heat map with colour range is scaled between the blue and yellow. The yellow side is positive tree-fallen feature caused by strong wind of typhoon.

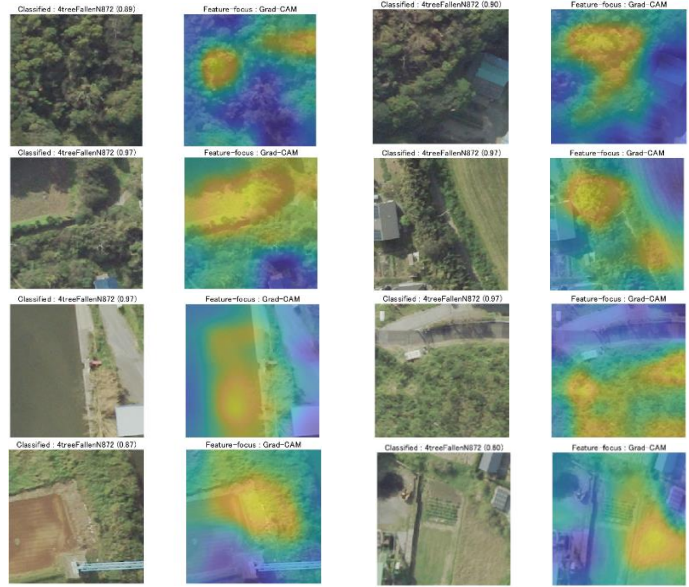

**Fig. 4.** Visual explanation of tree-fallen (yellow-blue range) using Grad-CAM, each pair of original clip and activation map. The yellow side is positive tree-fallen caused by strong wind.

## 4 Concluding Remarks

### 4.1 Disaster Features Visualization for Immediate Response Support

This paper proposed a method to predict the damaged area highlighted on the typhoon disaster feature such as tree-fallen using aero-photographs images. This method was able to classify land covers without damages and with ones. We trained the preferable deep network classifier with advantage of disaster class accuracy, the ShuffleNet compared with ten practical candidates of usable deep neural networks. Using target class probabilities per unit grid, we were able to visualize disaster features map to scale the colour range from blue to yellow. Furthermore, we were able to explain the disaster features on each unit grid images to compute the convolutional activation map using Grad-CAM. We demonstrated training and test studies applied to aero-photographs with the size of 14 thousands pixels by 15 thousands ones recorded at the south Chiba region after eighteen days at the post Typhoon Faxai, 27 to 28 September 2019.

We aware that our method may have two primary limitations. The first is a supervised learning approach based on the past experienced data. Natural disaster are rare events, so the data mining opportunity is limited, and results in the scarcity of disaster region of interest images. The second is based on the input images from a point of sky-view using an aero-photography. We should incorporate the ground-view data to be more useful for immediate disaster response. We should be comprehensive learning using another resources such as public surveillance camera, private mobile contents.

### 4.2 Future Works for Disaster Visual Mining and Learning Variations

This paper studied the one of disaster features of strong wind due to typhoon. We also focused on the damaged class of house roof break, and computed the feature map and explain the vinyl seat on the roof using Grad-CAM. These results have been shown as Fig.5 in Appendix because of limited space. Further, we try to carry out data mining towards another supervised sky view images though disaster event is rare, so we need to continue to collect aerial photographs at next coming typhoon with different damage features, for example flooded area and levee break point. Though it needs many annotation works to divide subgroup of clips, we can build a prototype of classifier and using the base model we can almost automate to classify subgroup of clips. Furthermore, we can try to learn and represent various disaster features and so it could be possible to make the visual explanation for immediate response support to faster recovery.

**Acknowledgements.** We gratefully acknowledge the constructive comments of the anonymous referees. Support was given by the Aero Asahi Co.Ltd. of Jun Miura, who provided us the aerial photographs recorded at the Chiba after the Typhoon Faxai. We thank Takuji Fukumoto and Shinichi Kuramoto for supporting us MATLAB resources.

## Appendix

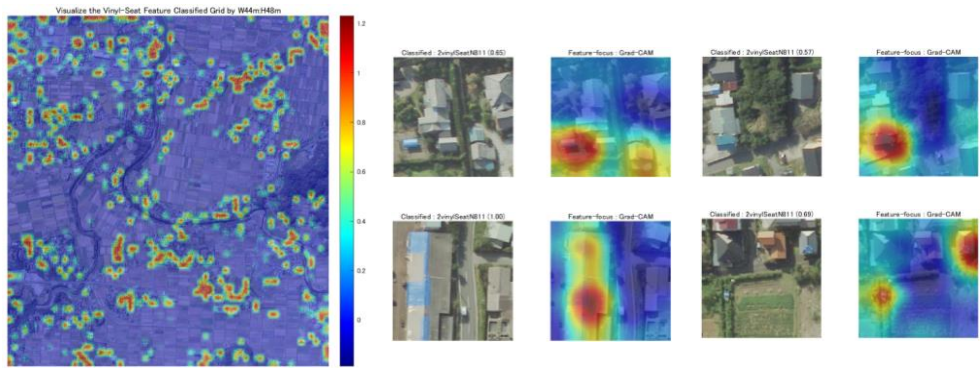

**Fig. 5.** (Left) House roof break feature map, (Right) Visual explanation of roof break (red-blue range) using Grad-CAM, each pair of original clip and activation map, roof break covered with vinyl seat. The red is positive roof break affected by strong wind.